\documentclass[final]{elsarticle}

\usepackage{amssymb}
\usepackage{longtable}
\usepackage{amsmath}
\usepackage{hyperref}
\usepackage{booktabs}
\usepackage{array,multirow}
\usepackage[utf8]{inputenc}
\usepackage[english]{babel}
\usepackage{longtable}
\usepackage{caption}
\usepackage{subcaption}
\usepackage{floatrow}
\usepackage[nolist]{acronym}
\usepackage{placeins}
\usepackage{algorithmicx}
\usepackage{algorithm}
\usepackage{algpseudocode}
\usepackage[load-configurations=version-1]{siunitx} 

\begin{document}

\def\*#1{\mathbf{#1}}
\newcommand{\specialcell}[2][c]{%
  \begin{tabular}[#1]{@{}c@{}}#2\end{tabular}}
\newcolumntype{H}{>{\setbox0=\hbox\bgroup}c<{\egroup}@{}}
\newcommand{\norm}[1]{\left\lVert#1\right\rVert}
\begin{acronym}[PCTKVM] 
    \acro{PCVM}[PCVM]{Probabilistic Classification Vector Machine}
    \acro{PCTKVM}[PCTKVM]{Probabilistic Classification Transfer Kernel Vector Machine}
    \acro{BT}[BT]{Basis-Transfer}
    \acro{TKL}[TKL]{Transfer Kernel Learning}
    \acro{TCA}[TCA]{Transfer Component Analysis}
    \acro{SVM}[SVM]{Support Vector Machine}
    \acro{JDA}[JDA]{Joint Distribution Adaptation}
    \acro{SVD}[SVD]{Singular Value Decomposition}
    \acro{EVD}{Eigen Value Decomposition}
    \acro{RKHS}[RKHS]{Reproducing Kernel Hilbert Space}
    \acro{PCA}[PCA]{Principal Component Analysis}
    \acro{NBT}[NBT]{Nyström Basis Transfer}
    \acro{SA}[SA]{Subspace Alignment}
    \acro{SVM}[SVM]{Support Vector Machine}
    \acro{NTVM}[NTVM]{Nyström Transfer Vector Machine}
    \acro{SURF}[SURF]{Speeded Up Robust Features Extraction}
\end{acronym}
\begin{frontmatter}

\title{Transfer Learning Extensions for the Probabilistic Classification Vector Machine \footnote{The final authenticated publication is available at \newline \url{https://doi.org/10.1016/j.neucom.2019.09.104}}}

\author[label1]{Christoph Raab\corref{cor1}}
\ead{christoph.raab@fhws.de}

\author[label1]{Frank-Michael Schleif}
\ead{frank.schleif@fhws.de}

\address[label1]{University for Applied Science Würzburg-Schweinfurt, Sanderheinrichsleitenweg 20, Würzburg, Germany}
\cortext[cor1]{I am corresponding author}

\begin{abstract}
    Transfer learning is focused on the reuse of supervised learning models in a new
    context. Prominent applications can be found in robotics, image processing or web mining. In these fields, the learning scenarios are naturally changing but often remain related to each other motivating the reuse of existing supervised models.
    Current transfer learning models are neither sparse nor interpretable.
    Sparsity is very desirable if the methods have to be used in technically limited environments and interpretability
    is getting more critical due to privacy regulations.
    In this work, we propose two transfer learning extensions integrated into the sparse and interpretable probabilistic classification vector machine.
    They are compared to standard benchmarks in the field and show their relevance either by sparsity or performance improvements.
\end{abstract}


\end{frontmatter}


\section{Introduction}\label{sec:introduction}
Supervised learning and in particular classification is an important task in machine learning with a  broad range of applications.
The obtained models are used to predict the label of unseen test samples.
In general, it is assumed that the underlying domain of interest is not changing between training and test samples.
If the domain is changing from one task to a related but different task, one would like to reuse the available learning model.
Domain differences are quite common in real-world scenarios and eventually lead to substantial performance drops \cite{Weiss2016}.

\begin{figure}[b!]
    \centering
   \begin{subfigure}[]{0.48\textwidth}
       \includegraphics[width=1\textwidth]{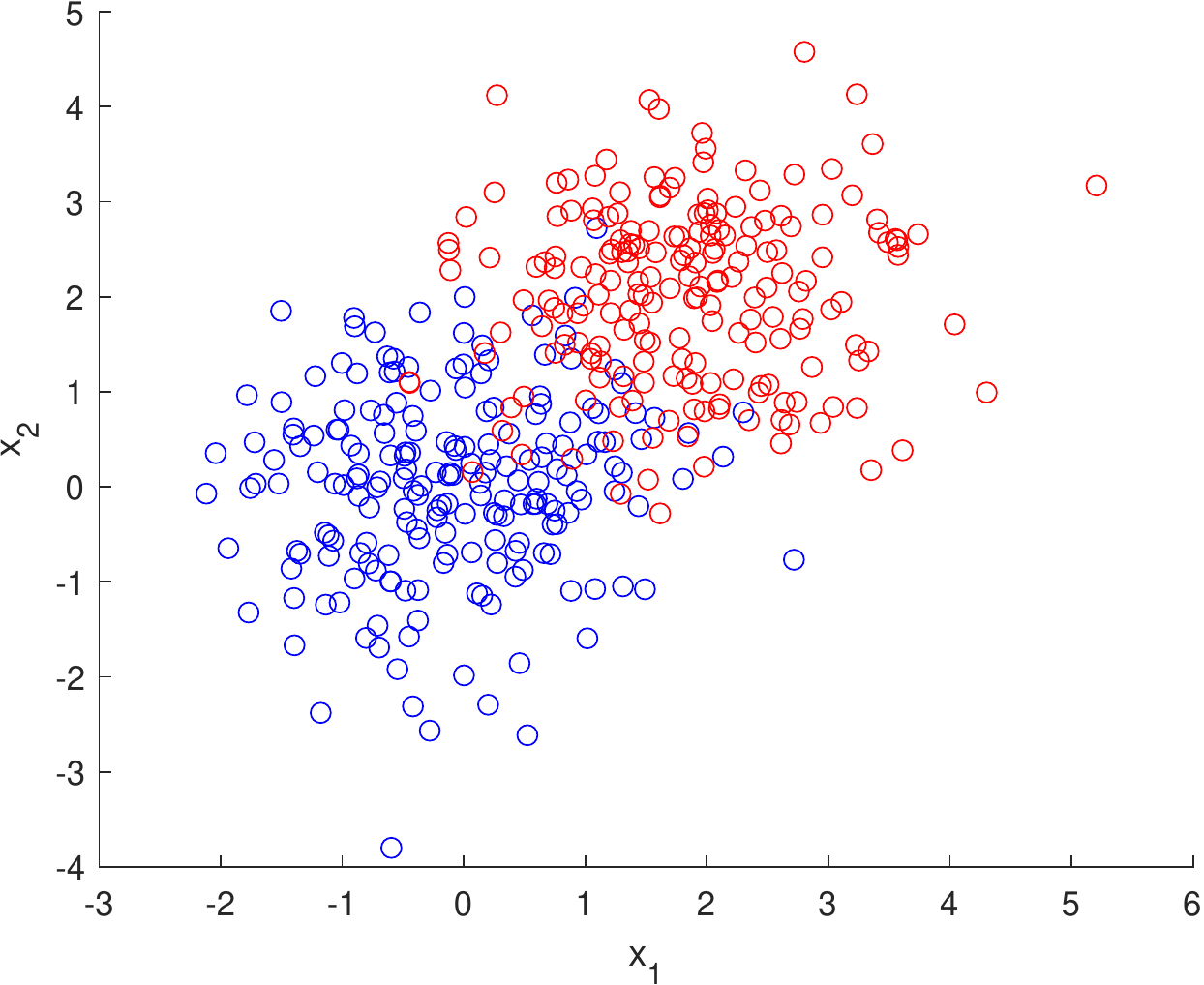}
       \subcaption{Traditional Problem}
       \label{FigTraditionalProblem}
   \end{subfigure}
   \begin{subfigure}[]{0.48\columnwidth}
    \includegraphics[width=1\textwidth]{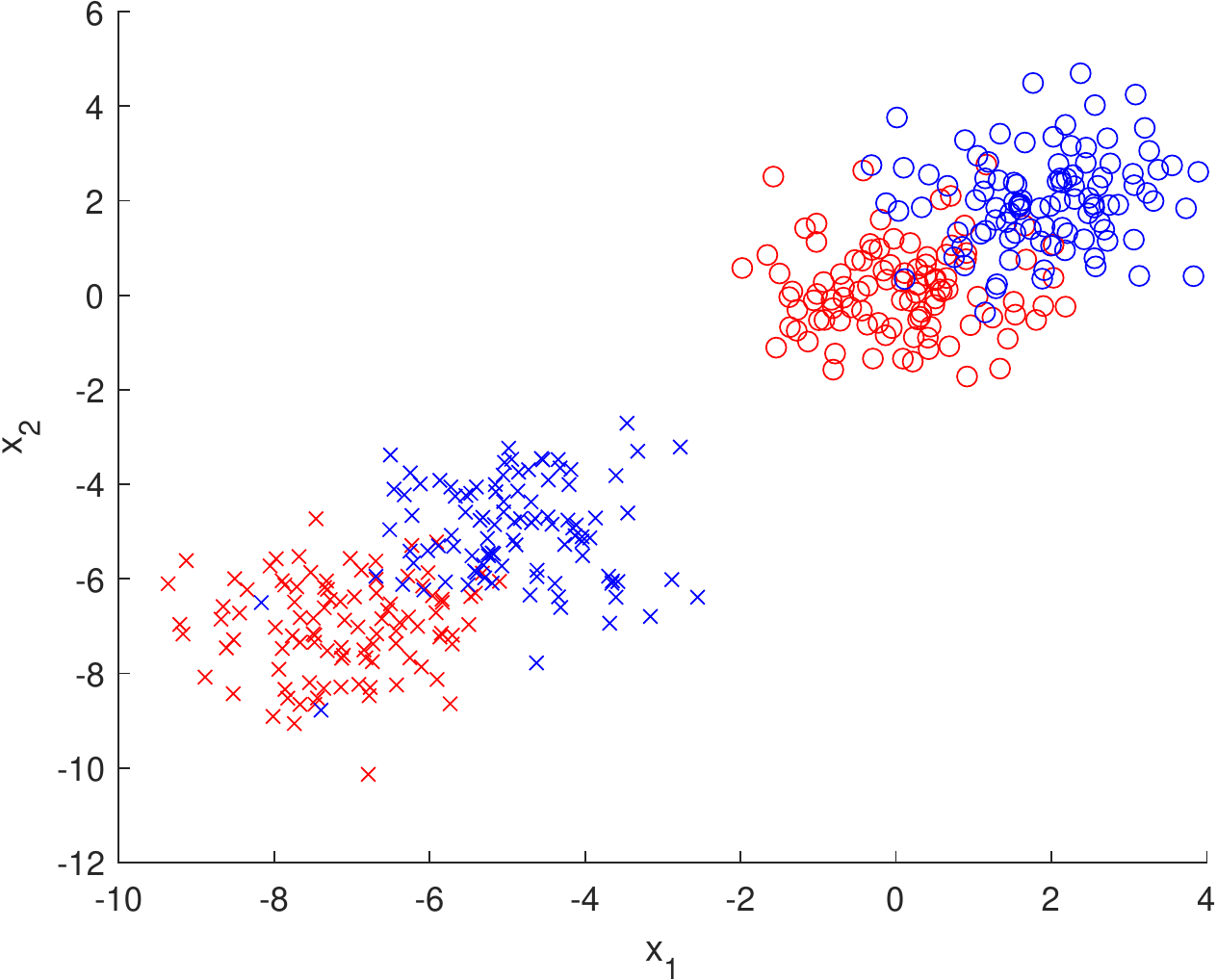}
    \subcaption{Transfer Problem}
    \label{FigTransferProblem}
    \end{subfigure}
   \caption{Toy example showing a comparison of a traditional classification task with two classes shown in red and blue. Left figure shows a traditional classification task with \textit{one domain} and right side shows transfer learning classification task with \textit{two domains}, which are indicated as shapes. Transfer learning aims to extract common information in one domain to help a model in another domain.\label{FigProblemComparison}}
\end{figure}

A transfer learning example is the classification of web pages:
A classifier is trained in the domain of university web pages with a word distribution according to universities and in the test scenario, the domain has changed to non-university web pages, where the word distribution may not be similar to training distribution.
Figure~\ref{FigProblemComparison} shows a toy example of a traditional and a transfer classification task with clearly visible domain differences.

More formally, let $\*{Z}=\{\*{z}_1,\dots,\*{z}_{n}\}$ be source data sampled from the source domain distribution $p(z)$ and let $\*{X}=\{\*{x}_1,\dots,\*{x}_{m}\}$ be target data from the target domain distribution  $p(x)$. Traditional machine learning assumes similar distributions, i.e.\ $p(z) \sim p(x)$, but transfer learning assumes different distributions, i.e.\ $p(z) \neq p(x)$, and appears in the web page example where $\*Z$ could be features of university websites and $\*X$ are features of non-university websites.

In general, transfer learning aims to solve the divergence between domain distributions by reusing information in one domain to help to learn a target prediction function in a different domain of interest \cite{5288526}. However, despite the definition, the proposed solutions implicitly solve differences by linear transformations, detailed in section~\ref{sec:sparsity_transfer} and~\ref{sec:performance_transfer}.
Multiple transfer learning methods have been already proposed, following different strategies and improving prediction performance of underlying classification algorithms in test scenarios \cite{Weiss2016}\cite{5288526}.
In this paper, we focus on sparse models, which are not yet covered sufficiently by transfer learning approaches.

The \ac{PCVM} \cite{Chen2009} is a sparse probabilistic kernel classifier pruning unused basis functions during training, found to be very effective
\cite{Chen2009}\cite{Schleif2015} with competitive performance to \ac{SVM} \cite{Cortes1995a}. The \ac{PCVM} is naturally sparse and creates interpretable models as needed in many application domains of transfer learning.
The original PCVM is not well suited for transfer learning due to their focus on stationary Gaussian distribution and is equipped within this work with two transfer learning approaches.

The contributions are detailed in the following:\newline
We integrate \ac{TKL} \cite{Long2015} into the \ac{PCVM} to retain its sparsity.
Inspired by the \ac{BT} \cite{stvm}, a subspace transfer learning approach is proposed and is also combined with \ac{PCVM}, boosting prediction performance significantly compared to the baseline. This is enhanced by Nyström techniques, which reduces computational complexity compared to \ac{BT}. Finally, a data augmentation strategy is proposed, making the approach independent of a certain domain adaptation task, which is a drawback of \ac{BT}.
The proposed solutions are tested against other commonly used transfer learning approaches on common datasets in the field.

The rest of the paper is organized as follows: An overview of related work is given in section~\ref{sec:relatedwork}. The mathematical preliminaries of \ac{PCVM} and Nyström approximation are introduced in section~\ref{sec:pre}. The proposed transfer learning extensions following in section~\ref{sec:ny_tl}. An experimental part is given in section~\ref{sec:experiments}, addressing the classification performance, the sparsity and the computational time of the approaches.
A summary and a discussion of open issues are provided in the conclusion.
\FloatBarrier
\section{Related Work}\label{sec:relatedwork}
The area of transfer learning provides a broad range of transfer strategies with many competitive approaches \cite{Weiss2016}\cite{5288526}.
In the following, we briefly name these strategies and discuss the key approaches used herein.

The \textit{instance transfer} methods try to align the distribution by re-weighting some source data, which can directly be used with target data in the training phase \cite{Weiss2016}.

Approaches implementing the \textit{symmetric feature transfer} \cite{Weiss2016} are trying to find a common latent subspace for source and target domain with the goal to reduce distribution differences, such that the underlying structure of the data is preserved in the subspace.
An example of a symmetric feature transfer method is the \ac{TCA} \cite{Pan2011}.

The \textit{asymmetric feature transfer} approaches try to transform source domain data in the target (subspace) domain.
This should be done in a way that the transformed source data will match the target distribution.
In comparison to the symmetric feature transfer approaches, there is no shared subspace available, but only the target space \cite{Weiss2016}.
An example is given by the \ac{JDA} \cite{Long2013}, which solves divergences in distributions similar to \ac{TCA}, but aligning conditional distributions with pseudo-labeling techniques.
Pseudo-labeling is performed by assigning labels to unlabeled target data by a baseline classifier, e.g. \ac{SVM}, resulting in a target conditional distribution, followed by matching it to the source conditional distribution of the ground truth source label \cite{Long2013}.
The \ac{SA} \cite{Fernando2013a} algorithm is another asymmetric transfer learning approach.
It computes a target subspace representation where source and target data are aligned but is only evaluated on image domain adaptation data.
We included \ac{SA} in the experimental study also containing non-image data.

The \textit{relational-knowledge transfer} aims to find some relationship between source and target data \cite{Weiss2016}.
\acf{TKL} \cite{Long2015} is a recent approach, which approximates a kernel of training data $K(\*{Z})$ with kernel of test data $K(\*{X})$ via the Nyström kernel approximation.
It only considers discrepancies in distributions and further claims it is sufficient to approximate a training kernel via test kernel, i.e.\ $K(\*{Z}) \approx K(\*{X})$, for effective knowledge transfer \cite{Long2015}. Note that the restriction to kernels does not apply to the Nyström transfer learning extension (section~\ref{sec:performance_transfer}) because it is completely in Euclidean space.

All the considered methods have approximately a complexity of $\mathcal{O}(n^2)$ where $n$ is the most significant number of samples concerning test or training \cite{Dai2007,Long2013,Long2015,Pan2011}.
According to the definition of transfer learning \cite{5288526}, these algorithms do \textit{transductive} transfer learning,
because  some \textit{unlabeled test} data \textit{must} be available at training time.
These transfer-solutions cannot be directly used as predictors, but instead are wrappers for classification algorithms. The baseline classifier is most often the \ac{SVM}. Note that discussed approaches are only tested with a classification task and may limited to this task.
\section{Preliminaries}\label{sec:pre}
\subsection{Probabilistic Classification Vector Machine}\label{sec:pcvm}
The \acf{PCVM} \cite{Chen2009} uses a probabilistic kernel 
model
\begin{equation}\label{pcvmModel}
f(\*{x};\*{w},b)=\Psi \left( \sum_{i=1}^{n} w_{i} \phi _{i}(\*{x})+b \right) = \Psi \left ( \Phi (\*{x})^\top \*{w} +b \right ) \in \mathbb{R}
\end{equation}
with a link function $\Psi(\cdot)$, $w_{i}$ being the weights of the basis functions $\phi _{i}(\*{x})$
and $b$ as bias term. The class assignment $c(\*x)$ of given data $\*x$ is given by \cite{Chen2009}
\begin{equation}
    c(x) = \begin{cases}
        1 & \Phi (\*{x})^\top \*{w} +b  \geq 0\\
        -1 & else.
    \end{cases}
\end{equation}
In \ac{PCVM} the basis functions $ \phi _{i}$ are defined explicitly as part of the model design.
In \eqref{pcvmModel} the standard kernel trick can be applied \cite{Scholkopf.2001}.
The probabilistic output of \ac{PCVM} \cite{Chen2009} is calculated by using the probit link function, i.e.\
\begin{equation}
\Psi (\*{x})=\int_{-\infty }^{x}\mathcal{N}(t|0,1)dt,
\end{equation}
where $\Psi (\*{x})$ is the cumulative distribution of the normal distribution $\mathcal{N}(0,1)$.
The \ac{PCVM} \cite{Chen2009} uses the \textit{Expectation-Maximization} algorithm for learning the model.
However, the \ac{PCVM} is not restricted to EM and other optimization approaches like Monte Carlo techniques are also possible \cite{Chen2014b}.
The underlying sparsity framework within the optimization prunes unused basis functions, independent of the optimization approach, and is, therefore, a sparse probabilistic learning machine.
In PCVM we will use the standard \textit{RBF}-kernel with a \textit{Gaussian} width $\theta$.

In \cite{Schleif2015} a PCVM with linear costs was suggested, which makes use of the Nyström approximation and could be additionally used to improve the run-time and memory complexity.
Further details can be found in \cite{Chen2009} and \cite{Schleif2015}.
\subsection{Nyström Approximation}\label{sec:ny_pre}
The computational complexity of calculating kernels or eigensystems scales with $\mathcal{O}(n^3)$ where $n$ is the sample size \cite{SchleifGT18}. Therefore, low-rank approximations and dimensionality reductions of data matrices are popular methods to speed up computational processes \cite{NIPS2000_1866}. The Nyström approximation \cite{NIPS2000_1866} is a reliable technique to approximate a kernel matrix $\*K$ by a low-rank representation, without computing the eigendecomposition of the whole matrix:\\
By the Mercer theorem, kernels $k(\*{x},\*{x^\prime})$ can be expanded by orthonormal eigenfunctions $f_i$ and non-negative eigenvalues $\lambda_i$ in the form
\begin{equation}
    k(\*{x},\*{x^\prime})=\sum_{i=1}^\infty \lambda_i f_i (\*{x}) f_i (\*{x^\prime}).
\end{equation}
The eigenfunctions and eigenvalues of a kernel are defined as solutions of
the integral equation
\begin{equation}
\int k(\*{x^\prime},\*{x}) f_i (\*{x}) p (\*{x}) d\*{x}
= \lambda_i f_i (\*{x^\prime}),
\end{equation}
where $p(\*{x})$ is a probability density over the input space.
This integral can be approximated based on the Nyström technique by an
i.i.d. sample $\{\*{x}_n\}_{k=1}^s$ from $p(\*{x})$
\begin{equation}
\frac{1}{s} \sum_{k=1}^s k(\*{x^\prime},\*{x}_n) f_i (\*{x}_n)
\approx \lambda_i f_i (\*{x^\prime}).
\label{eq:ep1}
\end{equation}
Using this approximation we denote with ${\*K}^{(s)}$ the corresponding
$s \times s$ Gram sub-matrix  and get the corresponding matrix eigenproblem equation as
\begin{equation}\label{eq:ny_sub_eigs}
    \frac{1}{s} {\*K}^{(s)} {\*U}^{(s)} = {\*U}^{(s)} \boldsymbol{\Lambda}^{(s)},
\end{equation}
with $ {\*U}^{(s)} \in \mathbb{R}^{s \times s}$ is column orthonormal and
$ \boldsymbol{\Lambda}^{(s)}$ is a diagonal matrix.
Now we can derive the approximations for the eigenfunctions and eigenvalues of the kernel $k$
\begin{equation}
\boldsymbol{\lambda_i} \approx \frac{\boldsymbol{\lambda}_i^{(s)} \cdot N}{s}, \quad  f_i (\*{x^\prime}) \approx \frac{\sqrt{s/N}}{ \boldsymbol{\lambda}_i^{(s)}} \*{k}_x^{\prime,\top} \*{u}_i^{(s)},
\label{eq:eigen-vec_func}
\end{equation}
where $\*{u}_i^{(s)}$ is the $i$th column of $\*{U}^{(s)}$.
Thus, we can approximate $f_i$ at an arbitrary point $\*{x^\prime}$ as long as we know the vector
$
\*{k}_x^\prime = (k(\*{x}_1,\*{x^\prime}), ... , k(\*{x}_s,\*{x^\prime})).
$
For a given $N \times N$ Gram matrix ${\*K}$ one may randomly chose $s$ rows and $s$ columns.
The corresponding indices are called landmarks, and should be chosen such that the
data distribution is sufficiently covered.
Strategies how to chose the landmarks have recently been addressed in \cite{DBLP:journals/tnn/ZhangK10a,Kumar.2012,GittensM16,DBLP:journals/csda/BrabanterBSM10}.
The approximation is exact if the sample size is equal to the rank of the original matrix and the rows of the sample matrix are linear independent.

\subsection{Nyström Matrix Form}
The technique just introduced can be simplified by rewriting it in matrix form \cite{Nemtsov2016}, where scaling factors like in equation~\eqref{eq:eigen-vec_func} are neglected. We will use the matrix formulation throughout the  remaining paper. Again, given a Gram  matrix $\*{K} \in \mathbb{R}^{n\times n}$, it can be decomposed to
\begin{equation}\label{eq:matrix_decomposition}
    \*{K}=
    \begin{bmatrix}
        \*{A} & \*{B} \\
        \*{C} & \*{D} \\
    \end{bmatrix},
\end{equation}
with $\*{A} \in \mathbb{R}^{s\times s}$, $\*{B} \in \mathbb{R}^{s\times (n-s)}$, $\*{C} \in \mathbb{R}^{(n-s)\times s} $ and $\*{D} \in \mathbb{R}^{(n-s)\times (n-s)}$.
The submatrix $\*{A}$ is called the landmark matrix containing $s$ randomly chosen rows and columns from $\*K$ and has the \ac{EVD} $\*{A}=\*{U}\boldsymbol{\Lambda}\*{U}^{-1}$ as in equation~\eqref{eq:ny_sub_eigs}, where eigenvectors are $\*{U}\in \mathbb{R}^{s\times s}$ and eigenvalues are on the diagonal of $\boldsymbol{\Lambda}\in \mathbb{R}^{s\times s}$.
The remaining approximated eigenvectors $\hat{\*U}$ of $\*K$, i.e.\ $\*C$ or $\*B^T$, are obtained by the Nyström method with $\hat{\*{U}}\boldsymbol{\Lambda}=\*{CU}$ as in equation~\eqref{eq:eigen-vec_func}.
Combining $\*U$ and $\hat{\*U}$ the \textit{full} approximated eigenvectors of $\*{K}$ are
\begin{equation}\label{eq:ny_eigenvector}
    \tilde{\*{U}} =
        \begin{bmatrix}
        \*U \\
        \hat{\*U}
    \end{bmatrix}
    =
    \begin{bmatrix}
        \*U \\
        \*C \*U \boldsymbol{\Lambda}^{-1}
    \end{bmatrix} \in \mathbb{R}^{n\times s}.
\end{equation}
The eigenvectors of $\*K$ can be inverted by computing (Note $\*C = \*B^{T}$):
\begin{equation}\label{eq:ny_left_eigenvector}
    \tilde{\*{V}}
    =\begin{bmatrix}
        \*{U}^{-1}&\boldsymbol{\Lambda}^{-1}\*{U}^{-1}\*{B}
    \end{bmatrix}.
\end{equation}
Combining equation~\eqref{eq:ny_eigenvector}, equation~\eqref{eq:ny_left_eigenvector} and $\boldsymbol{\Lambda}$, the matrix $\*{K}$ is approximated by
\begin{equation}\label{eq:ny_decomposition}
        \tilde{\*{K}}= \tilde{\*{U}} \boldsymbol{\Lambda} \tilde{\*{V}}=
        \begin{bmatrix}
            \*{U} \\
            \*{CU}\boldsymbol{\Lambda}^{-1}
        \end{bmatrix}
        \boldsymbol{\Lambda}
        \begin{bmatrix}
            \*{U}^{-1}& \boldsymbol{\Lambda}^{-1}\*{U}^{-1}\*{B}
        \end{bmatrix}.
\end{equation}
The Nyström approximation error is given by the Frobenius Norm between ground truth and reconstructed matrices, i.e.\ $error_{ny}=\norm{\tilde{\*{K}} - \*{K}}_F$.

\subsection{Kernel Approximation}\label{sec:kernel}
The Nyström approximation \cite{NIPS2000_1866} speeds up kernel computations, because kernel evaluation must not be done over all points, but only a fraction of it.
Let $\*{K} \in \mathbb{R}^{n\times n}$ be a Mercer kernel matrix with decomposition as in equation~\eqref{eq:matrix_decomposition}. Again we pick $s$ samples with $s \ll n$, leading to $\*A$ as defined before. An approximated kernel is constructed by combining equation~\eqref{eq:ny_decomposition} and \eqref{eq:matrix_decomposition}.
\begin{equation}\label{eq:nyst_kernel_approx}
    \tilde{\*K} =
    \begin{bmatrix}
        \*A  & \*B \\
        \*C & \*C \*A^{-1} \*B
    \end{bmatrix}
    =
    \begin{bmatrix}
        \*A \\
        \*C
    \end{bmatrix}
    \*A^{-1}
    \begin{bmatrix}
        \*A \*B
    \end{bmatrix} = \*K_{n,s} \*K_{s,s}^{-1} \*K_{s,n} ,
\end{equation}
with $\*K_{n,s}$ being a sub-matrix of $\*{K}$ incorporating all $n$ rows and $s$ landmark columns and $\*K_{s,s,}$ as landmarks matrix. Based on the definition of kernel matrices, $\*K_{s,n} =\*K_{n,s}^T$ is valid and, therefore, only $\*K_{n,s} $ and $ \*K_{s,s}$ must be computed. Note that the kernel approximation is used in section~\ref{sec:sparsity_transfer}.
\subsection{General Matrix Approximation}\label{sec:general}
Despite its initial restriction to kernel matrices, recent research expanded the Nyström technique to approximate a \ac{SVD} \cite{Nemtsov2016}.\\
Nyström-\ac{SVD} generalizes the concept of matrix decomposition with the consequence that respective matrices must not be square.\\
Let $\*G \in \mathbb{R}^{n\times d}$ be a rectangular matrix with decomposition as in equation~\eqref{eq:matrix_decomposition}.
The \ac{SVD} of the landmark matrix is given by $\*A = \*L \*S \*R^T$ where $\*L$ are left and $\*R$ are right singular vectors.
$\*S$ are positive singular values.
The decomposition matrices have the same size as in section~\ref{sec:ny_pre}.
Similarly to \ac{EVD} in section~\ref{sec:ny_pre} the left and right singular vectors for the non-symmetric part $\*C$ and $\*B$ are obtained via Nyström techniques \cite{Nemtsov2016} and are defined as $\hat{\*L} =  \*C \*R \*S^{-1}$ and $\hat{\*R} = \*B^T \*L \*S^{-1} $ respectively.\newline
Applying the same principals as for Nyström-\ac{EVD}, $\*G$ is approximated by
\begin{equation}\label{eq:ny_svd}
    \tilde{\*G} = \tilde{\*L} \*S \tilde{\*R}^T =
    \begin{bmatrix}
        \*L \\
        \hat{\*L}
    \end{bmatrix}
    \*S
    \begin{bmatrix}
        \*R &  \hat{\*R}
    \end{bmatrix}
    =
    \begin{bmatrix}
        \*L \\
        \*C \*R \*S^{-1}
    \end{bmatrix}
    \*S
    \begin{bmatrix}
        \*R &  \*S^{-1} \*L^T \*B
    \end{bmatrix}.
\end{equation}
Note that for non-gram matrices like $\*G$, $C=B^T$ is no longer valid. The matrix approximation (equation ~\eqref{eq:ny_svd}), which is described in this section, is used in the performance extension in section~\ref{sec:performance_transfer}.
\section{Nyström Transfer Learning}\label{sec:ny_tl}
\subsection{Transfer Kernel Sparsity-Extension}\label{sec:sparsity_transfer}
The domain invariant \ac{TKL} \cite{Long2015} technique is part of the first transfer learning extension.
Based on experimental results shown in section~\ref{sec:experiments} the use of \ac{TKL} retains model sparsity of \ac{PCVM}.
Given the categories of transfer learning \cite{Weiss2016} introduced in section~\ref{sec:relatedwork}, the approach can be seen as relational-knowledge-transfer.
It approximates the source kernel by the target kernel via the Nyström method.
It aims to find a source kernel close to target distribution and simultaneously searches an eigensystem of the source kernel, which minimizes the squared Frobenius norm between ground truth and approximated kernel \cite{Long2015}.

Let $\*{Z}=\{\*{z}_i\}_{i=1}^n \in \mathbb{R}^d$ be training data sampled from $p(z)$ with labels $\*Y_Z = \{y_i\}_{i=1}^n \in \mathcal{Y}=\{1,2,..,C\}$ and
$\*{X}=\{\*{x}_j\}_{j=1}^m \in \mathbb{R}^d$ be test data sampled from $p(x)$ with labels $\*Y_X = \{y_j\}_{j=1}^m \in \mathcal{Y}$. Note that we assume $p(z) \neq p(x)$. We obtain the associated kernel matrices $\*{K}_Z$ for training and $\*{K}_X$ for testing by evaluating an appropriate kernel function
(e.g.\ the RBF-kernel).
For clarity, we rewrite equation~\eqref{eq:matrix_decomposition} in kernel notation
\begin{equation}\label{EqTKLKernel}
\*{K}_A =
\begin{bmatrix}
\*{K}_X\>\>\>\>  \*{K}_{ZX} \\
\*{K}_{XZ} \>\>\>\> \*{K}_Z
\end{bmatrix},
\end{equation}
with $\*{K}_{Z}$ as train-, $\*{K}_{X}$ as test- and $\*{K}_{ZX}=\*{K}_{XZ}^T$ as \textit{cross-domain}-kernel between domains.
Revisiting the original Nystrom approach, it uses randomly chosen columns and rows from $\*K_A$, however, in \ac{TKL}, the target kernel $\*K_X$ is seen as a landmark matrix and is used for approximating the training kernel $\*K_Z$. Hence, landmarks are not randomly picked from $\*K_A$, but $\*K_X$ is used as landmark matrix and is the complete target set, therefore the approximation uses $m$ landmarks. $\*K_Z$ is not used in the landmark selection. This differs from the original Nyström approach.\newline
The \ac{TKL} approach assumes that the distributions differences are sufficiently aligned if $\*K_Z \simeq \*K_X$, which also leads to $p(K(\*Z)) \simeq p(K(\*X))$ \cite{Long2015}.\\
Rewriting equation~\eqref{eq:nyst_kernel_approx}, we can create an approximated training kernel by
\begin{equation}
    \*K_Z = \*U_Z \boldsymbol{\Lambda}_X \*U_Z^T = \*K_{ZX} \*K_X^{-1} \*K_{XZ},
\end{equation}
where $\*U_Z$ are eigenvalues from source kernel and $\boldsymbol{\Lambda}$ are eigenvalues from target kernel.
This new kernel is not based on the eigensystem of $\*K_X$ and should therefore not reduce distribution differences sufficiently.
Hence, the eigenvectors of $\*K_Z$ are constructed by the target eigensystem
\begin{equation}
    \bar{\*U}_Z = \*K_{ZX} \*U_X \boldsymbol{\Lambda}_X^{-1}.
\end{equation}
To fully approximate the eigensystem of the training kernel, the eigenvalues $\boldsymbol{\Lambda}=\{ \lambda_i\}_{i=1}^m$ are defined as model parameters of \ac{TKL}, leading to approximated kernel $\bar{\*K}_Z=\bar{\*U}_Z \boldsymbol{\Lambda} \bar{\*U}_Z^T$. These new parameters must be well-chosen to reduce domain differences and while keeping original training information.
The following optimization problem was suggested in \cite{Long2015} to solve this issue
\begin{equation}\label{eq:tkl_squared_frobenius}
    \begin{gathered}
    \min_{\boldsymbol{\Lambda}} || \bar{\*{K}}_{Z} - \*{K}_{Z}||^2_F =
    || \bar{\*{U}}_{Z} \boldsymbol{\Lambda}  \bar{\*{U}}_Z^T - \*{K}_Z ||^2_F \\
    \lambda_i \ge \zeta \lambda_{i+1}, i = 1,\dots,m-1 \quad
    \lambda_i \ge 0,  i = 0,\dots,m
    \end{gathered}
\end{equation}
with $\zeta \ge 1$ as eigenspectrum dumping factor. 
The obtained kernel is domain invariant \cite{Long2015} and can be used in any kernel machine.
The complexity of the \ac{TKL} algorithm can be given with $\mathcal{O}((d+r)(n+m)^2)$, where $r$ denotes the number of used eigenvectors and
$d$ refers to the dimensionality of data \cite{Long2015}. \ac{TKL} in combination with \ac{PCVM} is called \ac{PCTKVM}.
\subsubsection{Properties of PCTKVM algorithm}
The \ac{TKL} kernel $\bar{\*K}_Z=\bar{\*{U}}_{Z} \boldsymbol{\Lambda}  \bar{\*{U}}_Z^T$ is used to train the \ac{PCVM}.
In general, an RBF-kernel with in-place optimized distribution-width parameter $\theta$ is used in \ac{PCVM}.
Accordingly, a simple replacement of the standard RBF-kernel by a kernel obtained with \ac{TKL} will be inefficient.
In \ac{PCVM} the kernel is recalculated in each iteration, based on the optimized $\theta$ from the previous iteration.
Consequently, we would have to recalculate the entire transfer kernel too. The complexity of the standard \ac{PCVM}  is
$\mathcal{O}(l^3)$ where $l$ is the number of basis functions and $l=n$ at the beginning of the training and before pruning basis functions.
The complexity of \ac{TKL} is $\mathcal{O}((d+r)(n+m)^2)$. Combining them, we would end up with a computational complexity
of $\mathcal{O}(n^3(d+r)(n+m)^2)$.

However, the performance of \ac{PCTKVM} strongly depends on the quality of $\theta$. Hence, some reasonable $\theta$ must be obtained via grid search, because in-place optimization is infeasible. The \ac{PCTKVM} using a fixed $\theta$ has the complexity of $\mathcal{O}(n^3+m^2)$.
Note that \ac{TKL} and \ac{PCTKVM} are restricted to kernels, but are independent of the \textit{kernel type}.

Predictions are made with the \ac{PCVM} prediction function, but by employing $\bar{\*{K}}_{XZ} = \*{U}_X\*{\Lambda}\bar{\*{U}}_Z^T$ as kernel for test data \cite{Long2015}.
However, in case of the {\ac{SVM}} as baseline classifier, the kernel with size $n\times m$ is used and restricted to the respective support vectors.
The prediction function for the {\ac{SVM}} has the form $\hat{\*{y}} = \bar{\*{K}}_{ZX}(\*{\alpha} \cdot\*{y}_\*{Z})+b$, where $\*{\alpha}$ are the Lagrange multipliers \cite{Long2015}.

Because of the sparsity of the {\ac{PCVM}} the number of basis functions used in the decision function is typically small\footnote{The decision function may only be constructed by $10$ samples.}.
If we consider that our model has $l$ non-zero weight vectors with $l\ll n$ and because the {\ac{PCVM}} uses only kernel rows/columns corresponding to the non-zero weight vector index, our final kernel $\bar{\*{K}}_{LX}$ for prediction has size $(l\times m)$.
Therefore, the prediction function of the \ac{PCTKVM} has the form:
$
\hat{\*{y}} = \bar{\*{K}}_{LX}\*{w}+b.
$
The probabilistic output is calculated with the probit link function used in the {\ac{PCVM}}. Pseudo code of sparsity extension is shown in algorithm~\ref{PseudoCodePCTKVM}.
\begin{algorithm}
	\caption{Probabilistic Classification Transfer Kernel Vector Machine }\label{PseudoCodePCTKVM}
	\begin{algorithmic}[1]
		\Require $\*{K} = [\*{Z};\*{X}]$ as $N$ sized training and $M$ sized test set; $\*{Y}$ as $N$ sized training label vector; kernel(-type) \textit{ker}; eigenspectrum dumping factor $\zeta$; $\theta$ as kernel parameter.
		\Ensure Weight Vector $\*{w}$; bias $b$, kernel parameter $\theta$; transfer kernel $\bar{\*{K}}_\mathcal{A}$.
		\State $\*{D}$ = calculate\_dissimilarity\_matrix($\*{K}$);
		\State $\bar{\*{K}}_\mathcal{A}$ = transfer\_kernel\_Learning($\*{D}$,\textit{ker},$\theta$,$\zeta$); \Comment{According to equation~\eqref{eq:tkl_squared_frobenius}}
		\State [$\*{w}$,$b$] = pcvm\_training($\bar{\*{K}}_\mathcal{Z}$); \Comment{According to section~\ref{sec:pcvm}}
	\end{algorithmic}
\end{algorithm}
However, the sparsity extension performs similar to common approaches shown in the experiments in section~\ref{sec:experiments}, but to fully capture the advantages of Nyström and \ac{PCVM} the following section shows a transfer learning approach focused on performance.
\subsection{Nyström Basis Transfer Performance-Extension}\label{sec:performance_transfer}
It is a reasonable strategy in \ac{TKL} to align kernel matrices rather than kernel distributions in \ac{RKHS}, since distributions alignments are non-trivial in \ac{RKHS} \cite{Long2015}.
Hence, \ac{TKL} modifies the kernel explicitly to reduce the difference between two kernel matrices.
Similar source and target kernels must be obtained, because the underlying classifier is kernel-based and has no transfer learning.

\begin{figure}[]
     \centering
    \begin{subfigure}[]{0.5\textwidth}
        \includegraphics[width=0.95\textwidth]{images/Homogenoues_Transfer_Problem-eps-converted-to.pdf}
        \subcaption{Data unnormalized}
        \label{FigDataUnnorm}
    \end{subfigure}
    \begin{subfigure}[]{0.5\textwidth}
        \includegraphics[width=0.95\textwidth]{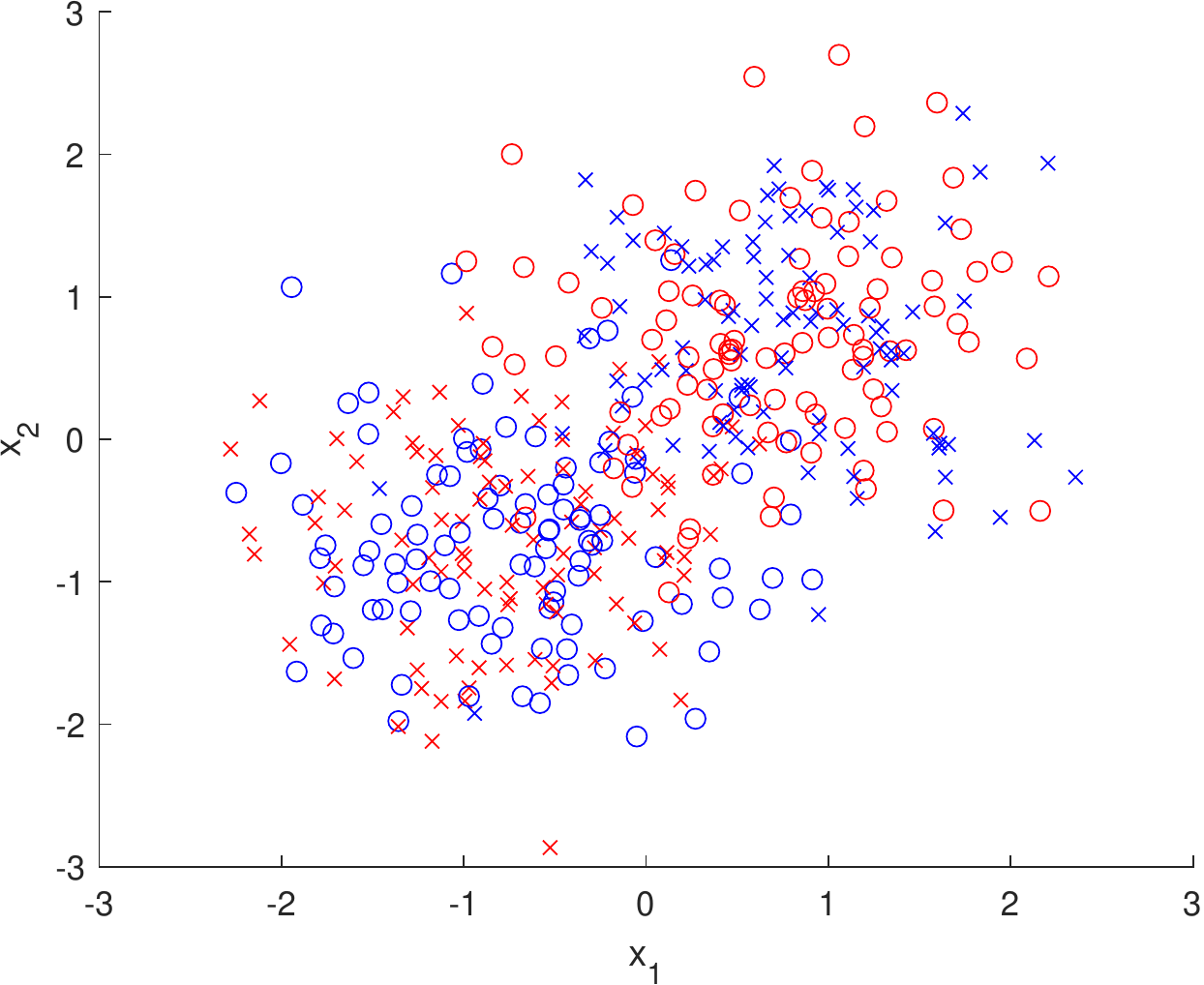}
        \subcaption{Data after standard normalization}
        \label{FigDataNorm}
    \end{subfigure}
    \begin{subfigure}[]{0.5\textwidth}
        \includegraphics[width=0.95\textwidth]{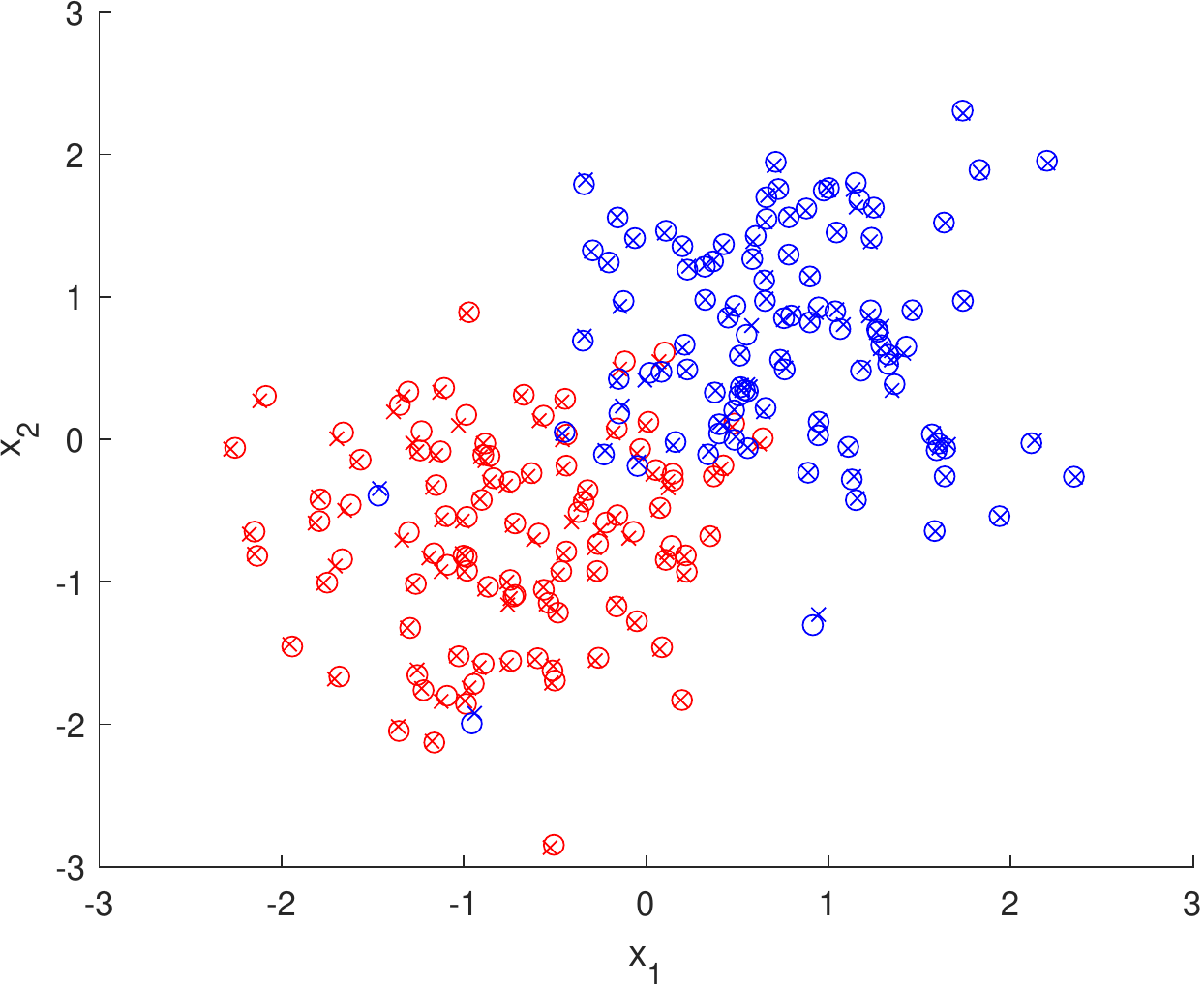}
        \subcaption{Data after Basis-Transfer}
        \label{FigBt}
\end{subfigure}
\caption{Process of \textit{Basis-Transfer} with samples from two domains. Class information is given by color (red/blue) and domain is indicated by shape (domain one - $o$, domain two - $\ast$). First (a), the non-normalized data with a knowledge gap. Second (b), a normalized feature space. Third (c), \textit{Basis-Transfer} approximation is applied, correcting the samples, i.e.\ shapes with same color are aligned, and training data is usable for learning a classification model. Best viewed in color.\label{FigBtProcess}}
\end{figure}

In \cite{6790375} is shown that if source and target datasets are similar, they follow similar distributions, i.e.\ if $\*Z \simeq \* X$ then $p(z)\simeq p(x)$, and further have similar kernel distributions and similar kernels.
Therefore, we propose a transfer learning approach operating in Euclidean space rather than \ac{RKHS}, because it does not limit approaches to kernel classifiers. Further, the obtained kernels after transfer of data also follow similar distributions.
A recent study \cite{stvm} already showed great transfer capabilities and performance, by aligning $\*X$ and $\*Z$ with a small error in terms of the Frobenius norm. However, this requires same samples sizes of $\*Z$ and $\*X$ and is assumed in the following with size $m\times d$.
The study considered the following optimization problem
\begin{equation}\label{eq:transfer_opt_problem}
    \min\limits_{\*M,\*T} \norm{ \*M\*Z\*T - \*X}_F^2
\end{equation}
where $\*M \in \mathbb{R}^{m\times m}$ and $\*T \in \mathbb{R}^{d\times d}$ are transformation matrices drawing the data closer together.
A solution is found analytically, summarized in three steps \cite{stvm}:
First, normalize data to standard mean and variance.
This will align marginal distributions in Euclidean space without considering label information \cite{stvm}.
Second, compute an \ac{SVD} of source and target data, i.e.\ $\*Z = \*L_Z \*S_Z \*R_Z^T$ and $\*X = \*L_X \*S_X \*R_X^T$.
Next, the approach assumes $\*S_Z \sim \*S_X$ in terms of Frobenius norm due to normalization with zero mean and variance one, reducing the scaling factor of singular values to the same range.
Finally, compute a solution for equation~\eqref{eq:transfer_opt_problem} by solving the linear equations.
One obtains $\*M = \*L_X\*L_Z^{-1}$ and $\*T = \*R_Z^{-1} \*R_X^T$.
Note that $\*L_Z\*L_Z^{-1} = \*R_Z\*R_Z^{-1} = \*I$.
Apply the transfer operation and approximate the source matrix by using target basis information
\begin{equation}
  \tilde{\*Z} = \*M \*Z \*T = \*L_X\*L_Z^{-1}  \*L_Z \*S_Z \*R_Z \*R_Z^{-1} \*R_X^T=\*L_X \*S_Z \*R_X^T,
\end{equation}
with $\tilde{\*Z} \in \mathbb{R}^{m \times d}$ as approximated source data, used for training.
The three-step process is shown in figure~\ref{FigBtProcess} as geometrical interpretation demonstrated with a toy example created by Gaussian random sampling.

In the following, the work \cite{stvm} is continued, and we propose a Nyström based version with three main improvements:
Reduction of computational complexity via Nyström, implicit dimensionality reduction and neglecting sample size requirements of \ac{BT}.
Further, we introduce a data augmentation strategy that eliminates the restriction \cite{stvm} to the task of text transfer learning.

Recap equation~\eqref{eq:transfer_opt_problem} and consider a slightly changed optimization problem
\begin{equation}\label{eq:transfer_opt_problem_ny}
    \min\limits_{\*M} || \*M\*Z - \*X||_F^2,
\end{equation}
where $\*Z, \*X \in \mathbb{R}^{m \times d}$, a transformation matrix $\*M \in \mathbb{R}^{m\times m}$ must be found, which is again obtained analytically.
Because we apply a dimensionality reduction technique, just the left-sided transformation matrix must be determined, which is derived in the following:
Based on the relationship between \ac{SVD} and \ac{EVD}, the \ac{PCA} can be rewritten in terms of \ac{SVD}. Consider the target matrix with \ac{SVD}
\begin{equation}
    \*X^T\*X = (\*R_X \*S_X \*L_X^T) (\*L_X \*S_X \*R_X^T) = \*R_X \*S_X^2 \*R_X^T,
\end{equation}
where $\*R_X \in \mathbb{R}^{d\times s}$ as eigenvectors and $\*S_X^2 \in \mathbb{R}^{s \times s}$ as eigenvalues of $\*X^T\*X$. By choosing only the biggest $s$ eigenvalues and corresponding eigenvectors the dimensionality of $\*X$ is reduced by
\begin{equation}\label{eq:proof_valid_pca}
    \*X_s = \*X \*R_s = \*L_s \*S_s \*R_s^T \*R_s = \*L_s \*S_s,
\end{equation}
where $\*L_s \in \mathbb{R}^{m\times s}$, $\*S_s \in \mathbb{R}^{s\times s}$ and $\*X_s \in \mathbb{R}^{m\times s}$ is the reduced target matrix.
Hence, only a left sided transformation in equation~\eqref{eq:transfer_opt_problem_ny} is required, because right sided transformation is omitted in equation~\eqref{eq:proof_valid_pca}.

The computational complexity of \ac{BT} and \ac{PCA} is decreased by applying Nyström-\ac{SVD}:
Let $\*Z$ and $\*X$ have a decomposition given as in equation~\eqref{eq:matrix_decomposition}.
Note for clarity the Nyström notation is used as in section~\ref{sec:general}.
For a Nyström-\ac{SVD} we chose from both matrices $s$ columns/rows obtaining landmarks matrices $\*A_Z = \*L_Z \*S_Z\*R_Z^T  \in \mathbb{R}^{s\times s}$  and $\*A_X = \*L_X \*S_X \*R_X^T \in \mathbb{R}^{s\times s}$.
Based on Nyström-\ac{SVD} in equation~\eqref{eq:ny_svd}, the dimensionality is reduced as in equation~\eqref{eq:proof_valid_pca} keeping only most relevant data structures
\begin{equation}\label{eq:ny_test_approx}
    \*X_s = \tilde{\*L}_X \*S_X =
    \begin{bmatrix}
        \*L_X \\
        \hat{\*L}_X
    \end{bmatrix}
    \*S_X
    =
    \begin{bmatrix}
        \*L_X \\
        \*C_X \*R_X \*S_X^{-1}
    \end{bmatrix} \*S_X \in \mathbb{R}^{m\times s},
\end{equation}
Hence, it is sufficient to compute an \ac{SVD} of $\*A_X$ instead of $\*X$ with $s\ll \{m,d\}$ and therefore is considerably lower in computational complexity.
Analogy, we approximate source data by $\*Z_s =  \tilde{\*L}_Z \*S_Z  \in \mathbb{R}^{n\times s}$.
Since we again assume $\*S_Z \sim \*S_Z$ due to data normalization, solving the linear equation as a possible solution for equation~\eqref{eq:transfer_opt_problem_ny}, leads to $\*M = \tilde{\*L}_X \tilde{\*L}_Z^{-1}$. Plugging it back we obtain
\begin{equation}\label{eq:ny_training_approx}
    \*Z_s = \tilde{\*L}_X \tilde{\*L}_Z^{-1}\tilde{\*L}_Z \*S_Z =\tilde{\*L}_X \*S_Z  \in \mathbb{R}^{m\times s}
\end{equation}
where again the basis of target data transfers structural information into the training domain.
The matrix $\*Z_s$ is used for training and $\*X_s$ is used for testing.
According to \cite{Weiss2016}, it is an asymmetric transfer approach. Further, it is transductive \cite{5288526} and does not need labeled target data.
For further references, we call the approach \ac{NBT} and in combination with \ac{PCVM}, \ac{NTVM}.
\subsubsection{Properties of Nyström Basis Transfer}\label{sec:props_nbt}
We showed that \ac{NBT} is a valid \ac{PCA} approximation by equation~\eqref{eq:proof_valid_pca}.
It follows by definition of \ac{SVD} that $\tilde{\*L}_Z\tilde{\*L}_Z^T = \tilde{\*L}_X\tilde{\*L}_X^{-1} =\*I$ and $\tilde{\*L}_X$ is an orthogonal basis.
Therefore, equation~\eqref{eq:ny_test_approx} and equation~\eqref{eq:ny_training_approx} are orthogonal transformations.
In particular equation~\eqref{eq:ny_training_approx} transforms the source data into the target subspace and projects it onto the principal components of $\*X$.
If data matrices $\*X$ and $\*Z$ are standard normalized\footnote{Experimental data are standard normalized to mean zero and variance one in the preprocessing.}, the geometric interpretation is a rotation of source data w.r.t to angles of the target basis, already shown in figure~\ref{FigBtProcess}.\\
The computational complexity of \ac{NBT} is given by the complexity of Nyström-\ac{SVD} and the calculation of the respective landmark matrices $\*A_Z$ and $\*A_X $ with complexity $\mathcal{O}(s^2)$.
The matrix inversion of diagonal matrix $\*S_X^{-1}$ in equation~\eqref{eq:ny_test_approx} can be neglected.
Remaining matrix multiplications are $\mathcal{O}(ns^2)$, contributing to an overall complexity of \ac{NBT}, which is $O(s^2)$ with $s\ll \{n,m\}$. This makes \ac{NBT} the fastest transfer learning solution in terms of computational complexity in comparison to the discussed methods in section~\ref{sec:relatedwork}.\newline
The approximation error is similar to the original Basis-Transfer \cite{stvm} error:
\begin{equation}
    error_{nbt}=\norm{\*X_s - \*Z_s}_F = \norm{\tilde{\*L}_X \*S_X - \tilde{\*L}_X \*S_Z}_F = \norm{\*S_X - \*S_Z}_F
\end{equation}
\subsubsection{Data Augmentation}
In \ac{BT} \cite{stvm}, sample sizes of data matrices must be aligned.
This is not required in \ac{NBT} as seen in equation~\eqref{eq:ny_test_approx} and equation~\eqref{eq:ny_training_approx}, because differences in size are aligned during transformation.
However, the original dataset $\*Z$ has an $n\times 1$ sized label vector with $n\neq m$, which does not correspond to $\*Z_s$ and this label-vector should not be transformed into the new size because semantic label information does not correspond with transformed data.
Hence, sample sizes must still be the same, i.e.\ $m=n$, but is not required by definition of \ac{NBT}.
We propose a data augmentation strategy for solving different sample sizes, \textit{applied before} doing knowledge transfer.
Data augmentation is common in machine or deep learning and has a variety of applications \cite{Zhang2016,Hauberg16}.
However, source and target data should have a reasonable size to proper encode domain knowledge.

In general, there are two cases, first $n < m$, meaning there is not enough source data. This is augmented via sampling from a class-wise multivariate Gaussian distribution $\mathcal{N}_d$ harmonizing the number of samples per class of source data.
The other case is $n>m$ and is solved by uniform random removal of source data from the largest class, i.e.\ $\mathcal{U}(1,|c_{max}|)$ with $|c_{max}|$ as number of class samples and $c_{max} = \{c_i \mid \forall c_i \>\> max(|c_i|)\}$ as label with most samples, in the source set $(\*Z,\*Y_Z)$. The approach reduces source data to size $m$.
This is somewhat counter-intuitive because one does not want to reduce the source set.
However, we have no class information of the target set at training time, and we would be guessing class labels of target data when adding new artificial samples.
The data augmentation strategy is summarized as
\begin{equation}\label{eq:data_augmentation}
    \{\*Z,\*Y_Z\} =
    \begin{cases}
        \{\*Z,\*Y_Z\} \cup
        \{ \*z = \mathcal{N}_d (\boldsymbol{\mu}_{c_i},\boldsymbol{\sigma}_{c_i}),  y = \{c_i \mid \forall c_i \>\> min(|c_i|)  \} \} &  n < m\\
        \{\*Z,\*Y_Z\} \setminus \{ \*z, y | \*z \in \*Z \cap p(\*z) = \frac{1}{|c_{max}|} \cap y(\*z_i)=c_{max}  \} &  m < n
    \end{cases}
\end{equation}
where $\boldsymbol{\mu}_{c_i}= \frac{1}{|c_i|}\sum_{y(\*z_i) = c_i} \*z_i$ is class-wise mean, $\boldsymbol{\sigma}_{c_i} = \frac{1}{|c_i|}\sum_{y(\*z_i) = c_i} (\*z_i - \boldsymbol{\mu}_{c_i})^2$ is class-wise variance.  The function $y(\cdot)$ maps a training sample to the ground truth label $c_i \in \mathcal{Y}$ and $|c_i|$ is the number of class sample occurrences.

Pseudo code of Data Augmentation, \ac{NBT} and \ac{PCVM} summarized as \ac{NTVM} is shown in algorithm~\ref{alg:pseudocodentm}.

\begin{algorithm}
    \caption{Nyström Transfer Vector Machine (\ac{NTVM})}\label{alg:pseudocodentm}
    \begin{algorithmic}[1]
        \Require $\*{Z}$ as $n$ sized training; $\*X$ as $m$ sized test set; $\*{Y}$ as $n$ sized training label vector; $s$ as number of landmarks parameter; $\theta$ as RBF-kernel parameter.
        \Ensure Weight Vector $\*{w}$; bias $b$;
        \State $[ \*Z_n,\*X_n ]$=standard\_normalization($\*Z$,$\*X$) \Comment{Similar as in Fig.~\ref{FigDataNorm}}
        \State $ \*Z_a,\*Y_a $ = data\_augmentation($\*Z_n$,$\*Y$)                \Comment{According to equation~\eqref{eq:data_augmentation}}
        \State $\*A_Z$ = matrix\_decomposition($\*Z_a$,$s$) \Comment{According to equation~\eqref{eq:matrix_decomposition}}
        \State $\*A_X,\*C_X$ = matrix\_decomposition($\*X_n$,$s$)   \Comment{According to equation~\eqref{eq:matrix_decomposition}}
        \State  $ \*S_Z$ = $SVD(\*{A}_Z)$; \Comment{Singular Values of landmark matrix of $\*Z$}
        \State  $ \*L_X, \*S_X, \*R_X$ = $SVD(\*{A}_X)$; \Comment{SVD of landmark matrix of $\*X$}
        \State   $\tilde{\*L}_X$ =  $\begin{bmatrix}
        \*L_X &
        \*C_X \*R_X \*S_X^{-1}
    \end{bmatrix}^T$ \Comment{According to equation~\eqref{eq:ny_test_approx}}
        \State $\*X_s = \tilde{\*L}_X \*S_X $ \Comment{According to equation~\eqref{eq:ny_test_approx}}
        \State $\*Z_s = \tilde{\*L}_X \*S_Z $ \Comment{According to equation~\eqref{eq:ny_training_approx} and similar to Fig~\ref{FigBt}}
        \State [$\*{w}$,$b$] = pcvm\_training($\*{Z}_{s}$,$\*{Y}$,$\theta$); \Comment{According to \cite{Chen2009}}
    \end{algorithmic}
\end{algorithm}
\section{Experiments}\label{sec:experiments}
We follow the experimental design typical for transfer learning algorithms \cite{Long2015,Gong2017,Long2013,5288526,Pan2011}.
A crucial characteristic of datasets for transfer learning is that domains for training and testing are different but related.
This relation exists because the train and test classes have the same top category or source. The classes themselves are subcategories or subsets.
The parameters for respective methods\footnote{Source code, parameters and datasets obtainable via \url{https://github.com/ChristophRaab/ntvm}} are determined for best performance in terms of accuracy via grid search evaluated on source data.

\subsection{Dataset Description}\label{sec:description}
The study consists of $24$ benchmark datasets and are already preprocessed. Reuters from \cite{WenyuanDai.2007}, 20-Newsgroup from \cite{Long2014a} and Caltech-Office from \cite{Gong2017}.
A summary of image and text datasets is shown in table~\ref{TableImgData} and table ~\ref{TableTextData}.
Respective datasets are detailed in the following.
\subsubsection{Image Datatsets}
\textbf{Caltech-256\footnote{\url{https://people.eecs.berkeley.edu/\~jhoffman/domainadapt/\#datasets_code}}- Office}:
The first, Caltech (\textit{C}) is an extensive dataset of images and initially contains of 30607 images within 257 categories. However, in this setting, only 1123 images are used to be related to the Office dataset. We adopt the sampling scheme from \cite{Gong2017}.
The Office dataset is a collection of images drawn from three sources, which are from \textit{Amazon (A)}, digital SLR camera \textit{DSLR (D)} and \textit{webcam (W)}. They vary regarding camera, light situation and size, but having 31 object categories, e.g.\ computer or printer, in common.
Duplicates are removed, as well as images, which have more than 15 similar Scale Invariant Feature Transform (SIFT) in common.\\
To get an overall collection of the four image sets, which are considered as domains, categories with the same description are taken.
From the Caltech and Office dataset, ten similar categories are extracted:
backpack, touring-bike, calculator, head-phones, computer-keyboard, laptop-101, computer-monitor, computer mouse, coffee-mug, and projector.
They are the class labels from one to ten.\\
With this, a classifier should be trained in the training domain, e.g.\ on projector images (Class One) from amazon (Domain A), and should be able to classify the test image to the corresponding image category, e.g.\ projector (Class One) images from Caltech (Domain C) against other image types like head-phones (Class Two).
The final feature extraction is done with \ac{SURF} and encoded with 800-bin histograms.
Finally, the twelve combination of domain datasets are designed to be trained and tested against each other by the ten labels \cite{Gong2017}. An overview of the image dataset is given in table~\ref{TableImgData}.
\begin{table}[t]
    \centering
    \resizebox{\textwidth}{!}{%
    \begin{tabular}{ccccc}
        \toprule
        Dataset & Subsets   & \#Samples & \#Features & \#Classes  \\\midrule
        Caltech-256  & Caltech (C) & 1123      &  \multirow{4}{*}{800}           &   \multirow{4}{*}{10}               \\
        Office  & Amazon (A)  & 958         &            &         \\
                &  DSLR (D) &   295       &            &         \\
                & Webcam(W)   &     157     &            &         \\\bottomrule
    \end{tabular}}
    \caption{Dataset characteristics of image domain adaptation datasets containing datasets and corresponding subsets, numbers of samples, features and classes.}
    \label{TableImgData}
\end{table}

\subsubsection{Text Datasets}
In the following, the text datasets are discussed. The arranging of the text domain adaption datasets is different from the image datasets. The text datasets are structured into top categories and subcategories. These top categories are regarded as labels and the subcategories are used for training and testing. The variation of subcategories between training and testing creates a transfer problem. The difference to image datasets is that at the image datasets the (sub) -categories are labels and the difference in the top category (source of images) between training and testing, e.g\ \textit{Caltech} to \textit{Amazon}, creates the transfer problem. An overview of the text datasets is given in table~\ref{TableTextData}.

\textbf{Reuters-21578}\footnote{\url{http://www.daviddlewis.com/resources/testcollections/reuters21578}}:
A collection of Reuters news-wire articles collected in 1987 with a hierarchical structure given as top-categories and subcategories to organize the articles. The three top categories \textit{Organization (Orgs)}, \textit{Places} and \textit{People} are used in our experiments.  The category \textit{Orgs} has 56 subcategories, \textit{Places} has 176 and \textit{People} has 269. In the category \textit{Places}, all articles about USA are removed making the three nearly even distributed in terms of articles.

We follow the sampling scheme from \cite{WenyuanDai.2007}, which will be discussed in the following.
Note that the top categories, which are just mentioned, are the labels of the datasets.

All subcategories of a top category are randomly divided into two parts of subcategories with about the same number of articles. This selection is fixed for all experiments.
For a top category $O$ this creates two parts $o1$ and $o2$ of subcategories and for another top category $P$ this creates the parts $p1$ and $p2$. The top category $O$ is regarded as a class and $P$ as another one.

The transfer problem is created by using $o1$ and $p1$ for training and $o2$ and $p2$ is used for testing. This is a two class problem, because of the two top categories $O$ and $P$. Such a configuration is called \textit{$O$ vs $P$}. If the second part is used for training, i.e.\ $o2$ and $p2$, and the first part for testing, i.e.\ $o1$ and $p1$, it is regarded as \textit{$P$ vs $O$}. The individual subcategories have different distribution but are related by the top category, creating a change in distribution between training and testing.

Based on this six datasets are generated: \textit{Orgs vs. Places}, \textit{Orgs vs. People}, \textit{People vs. Places}, \textit{Places vs. Orgs}, \textit{People vs. Places} and \textit{Places vs. People}.
The articles are converted to lower case, words are stemmed and stopwords are removed.
With the Document Frequency (DF)-Threshold of 3, the numbers of features are cut down.
The features are generated with Term-Frequency Inverse-Document-Frequency (TFIDF). For a detailed choice of subcategories see \cite{WenyuanDai.2007}.

\begin{table}[t!]
    \centering
    \resizebox{\textwidth}{!}{%
    \begin{tabular}{ccccc}
        \toprule
        Top Category (Names)  & Subcategory & \#Samples & \#Features & \#Classes \\\midrule
        Comp   & comp.graphics              & 970   &   \multirow{16}{*}{25804}          &   \multirow{16}{*}{2} \\
               & comp.os.ms-windows.misc    & 963   &                                   &    \\
               & comp.sys.ibm.pc.hardware   & 979   &                                   &    \\
               & comp.sys.mac.hardware      & 958   &                                   &    \\
        Rec    & rec.autos                  & 987   &                                   &    \\
               & rec.motorcycles            & 993   &            &      \\
               & rec.sport.baseball         & 991   &            &      \\
               & rec.sport.hokey            & 997   &            &      \\
        Sci    & sci.crypt                  & 989   &            &      \\
               & sci.electronics            & 984   &            &      \\
               & sci.med                    & 987   &            &      \\
               & sci.space                  & 985   &            &      \\
        Talk   & talk.politics.guns         & 909   &            &      \\
               & talk.politics.mideast      & 940   &            &      \\
               & talk.politics.misc         & 774   &            &      \\
               & talk.religion.misc         & 627   &            &      \\\midrule
        Orgs   & 56 subcategories          & 1237  & \multirow{3}{*}{4771}           & \multirow{3}{*}{2}           \\
        People & 269 subcategories          & 1208  &            &          \\
    Places & 176 subcategories              & 1016  &            &      \\\bottomrule
    \end{tabular}   }
    \caption{Dataset characteristics of text domain adaptation datasets containing top-categories and corresponding subcategories, numbers of samples, features and labels. Horizontal line separates dataset in 20-Newsgroup (upper half) and Reuters (lower half) \cite{Long2015}. At reuters there are many subcategories, therefore we only show the number of subcategories.}
    \label{TableTextData}
\end{table}

\textbf{20-Newsgroup}\footnote{\url{http://qwone.com/~jason/20Newsgroups/}}: The original collection has approximately 20000 text documents from 20 newsgroups. The four top categories are \textit{comp}, \textit{rec}, \textit{talk} and \textit{sci} and containing four subcategories each. We follow a data sampling scheme introduced by \cite{Long2015} and generate 216 cross domain datasets based on subcategories:
Again the top categories are the labels and the sub categories are varied between training and testing to create a transfer problem.

Let $C$ be a top category and $C1,C2,C3,C4$ are subcategories of $C$ and another top category with $K$ and $K1,K2,K3,K4$ are subcategories of $K$. A dataset is constructed by selecting two subcategories for each top-category, e.g.\ $C1$, $C2$, $K1$, and $K2$, for training and select another four, e.g.\ $C3$, $C4$, $K3$, and $K4$ for testing. The top categories $C$ and $K$ are respective classes.

For two top categories every permutation is used and therefore $C^2_4 \cdot K^2_4 = 36$ datasets are generated. By combining each top category with each other there are 216 dataset combinations. The datasets are summarized as mean per top category combination, e.g.\ \textit{C vs K}, which are \textit{comp vs rec}, \textit{comp vs talk}, \textit{comp vs sci}, \textit{rec vs sci}, \textit{rec vs talk} and \textit{sci vs talk}. The transfer problem is created by training and testing on different subcategories analogy to Reuters.
This version of \textit{20-Newsgroup} has 25804 TF-IDF features within 15033 documents \cite{Long2015}.


Note to reproduce the results below, one should use the linked version of datasets with same choice of subcategories.
Regardless of dataset, features have been normalized to standard mean and variance.
The samples for training and testing the classifiers are drawn with $5\times 2$-fold sampling scheme suggested by \cite{Alpaydm.1999}, with a transfer learning adapted data sampling scheme as suggested in \cite{Gong2017}.

\FloatBarrier
\subsection{Comparison of Prediction Performance}\label{sec:results_performance}
The results of the experiments are summarized in table~\ref{tab:result_error} and showing mean errors of the cross-validation study per dataset.
To determine statistically significant differences, we follow \cite{Chen2009}, using the Friedman Test \cite{Demsar2006} with a confidence level of $5\%$ and Bonferroni-Dunn Post-Hoc correction.
The $\ast$ marks statistical significance against \ac{NTVM}.
The \ac{PCTKVM} and \ac{NTVM} are compared to baseline classifier to standard transfer learning methods and \textit{non-transfer learning} baseline methods, i.e.\ \ac{SVM} and \ac{PCVM}.
The \ac{PCTKVM} has overall comparable performance to \ac{PCVM}, however, is worse at Newsgroup, showing negative transfer \cite{Weiss2016}. This should be investigated in future work.

The \ac{NTVM} method has excellent performance and outperforms every other algorithm by far.
In the overall comparison, the \ac{NTVM} is significantly better
compared to the other approaches, except \ac{SA}.

\begin{table}[h!]
    \resizebox{\textwidth}{!}{%
    \begin{tabular}{ccccccccc}
        \toprule
     Dataset       & SVM   & PCVM  & TCA   & JDA    & TKL   & SA   & \specialcell{PCTKVM \\ Our Work} & \specialcell{NTVM \\Our Work} \\\midrule
    Orgs vs People  & 23.0&31.1&23.8&25.7&18.6&6.3&20.7&\textbf{3.1}\\
    People vs Orgs  & 21.1&28.1&20.3&24.9&13.0&5.7&23.2&\textbf{3.1}\\
    Orgs vs Place   & 30.8&33.4&28.7&27.4&22.7&6.4&33.9&\textbf{2.6}\\
    Place vs Orgs   & 35.8&35.7&34.1&34.1&17.5&6.6&40.6&\textbf{2.4}\\
    People vs Place & 38.8&41.3&37.2&41.3&30.6&8.3&36.3&\textbf{2.6}\\
    Place vs People & 41.3&41.2&43.4&43.3&34.0&11.5&39.7&\textbf{2.5}\\\midrule
    Reuters Mean    & 31.8 $\ast$ &35.1 $\ast$&31.3 $\ast$&32.8  $\ast$&22.7&7.5&32.4 $\ast$&\textbf{2.7}\\\midrule
    Comp vs Rec     & 12.7&17.9&8.1&7.8&3.0&1.8&37.3&\textbf{0.5}\\
    Comp vs Sci     & 24.5&29.1&26.3&27.1&9.5&4.8&33.7&\textbf{0.8}\\
    Comp vs Talk    & 5.1&6.2&2.9&4.2&2.4&\textbf{0.9}&14.9&7.2\\
    Rec vs Sci      & 23.7&36.6&17.3&23.9&5.1&1.6&29.3&\textbf{0.3}\\
    Rec vs Talk     & 18.7&27.8&13.6&15.2&5.6&\textbf{1.8}&33.4&3.1\\
    Sci vs Talk     & 21.7&30.9&20.1&26.1&14.6&\textbf{2.9}&30.2&6.9\\\midrule
    Newsgroup Mean  &17.8  $\ast$&24.7 $\ast$&14.7&17.4  $\ast$&6.7&\textbf{2.3}&29.8 $\ast$&3.1\\\midrule
    C vs A &48.0&55.9&49.7&49.0&49.1&38.0&52.8&\textbf{19.9} \\
    C vs W &53.8&57.0&55.1&53.5&\textbf{53.1}&68.1&55.2&59.5 \\
    C vs D &62.2&65.9&65.2&58.9&59.4&68.3&60.9&\textbf{47.4} \\
    A vs C &54.8&59.5&53.8&54.6&53.9&42.4&56.6&\textbf{37.5} \\
    A vs W &61.0&66.1&58.6&58.8&57.2&67.5&59.7&\textbf{55.6} \\
    A vs D &62.6&64.8&66.5&59.2&59.2&64.9&61.6&\textbf{48.9} \\
    D vs C &68.6&72.5&62.4&61.1&65.2&66.1&68.7&\textbf{19.9} \\
    D vs A &69.0&72.4&65.5&65.8&64.6&66.2&67.0&\textbf{35.6} \\
    D vs W &40.6&62.5&28.2&30.1&\textbf{27.4}&30.7&45.8&47.9 \\
    W vs C &65.9&67.8&62.1&65.3&63.2&59.4&63.9&\textbf{16.1} \\
    W vs A &67.6&69.0&63.3&67.0&63.0&64.7&68.4&\textbf{38.5} \\
    W vs C &23.1&41.1&\textbf{21.6}&22.1&27.4&32.5&45.5&59.8 \\ \midrule
    Image Mean & 56.4 $\ast$&62.9 $\ast$&54.3&53.8&53.6 &55.7 $\ast$&58.8 $\ast$&\textbf{40.6}\\ \midrule
    Overall Mean & 35.3  $\ast$&40.9  $\ast$&33.4  $\ast$&34.6 $\ast$&27.7 $\ast$&21.8&40.4 $\ast$&\textbf{15.5}\\
    \end{tabular}}
    \caption{Result of cross-validation test shown in mean error per dataset. Mean over dataset group at the end of each section. Bold marks winner. $\ast$ marks statistical differences with significance level of $0.05$ against \ac{NTVM}. The study shows that none of the listed algorithms is statistically significantly better as \ac{NTVM}.   \label{tab:result_error}}
\end{table}

Especially at Reuters, \ac{NTVM} convinces with stable and best performance over multiple datasets.
Table~\ref{tab:result_error} shows that \ac{NTVM} is significantly better in terms of mean at Reuters except \ac{TKL} and \ac{SA}.

The \ac{NTVM} also outperforms most of the time at image datasets, showing the capability to tackle multi-class problems and their independence from a certain domain adaptation task, unlike previous work \cite{stvm}.
Further, in terms of mean error on image, the \ac{NTVM} outperforms \ac{SVM}, \ac{PCVM}, \ac{PCTKVM} and \ac{SA} with statistical significant differences.

The \ac{NTVM} is also very good at Newsgroup, but  not that outstanding.
It is overall little worse than \ac{SA}, but not statistically significant.
Further, it is best at half of the datasets and convinces with error performances under one percent.

Note that the standard deviation is not shown, because it will not provide more insights into the performance. It is overall very similar and small, because the underlying classifier is the same.

\begin{figure}[t]
    \begin{subfigure}[c]{0.49\textwidth}

     \includegraphics[width=1\textwidth]{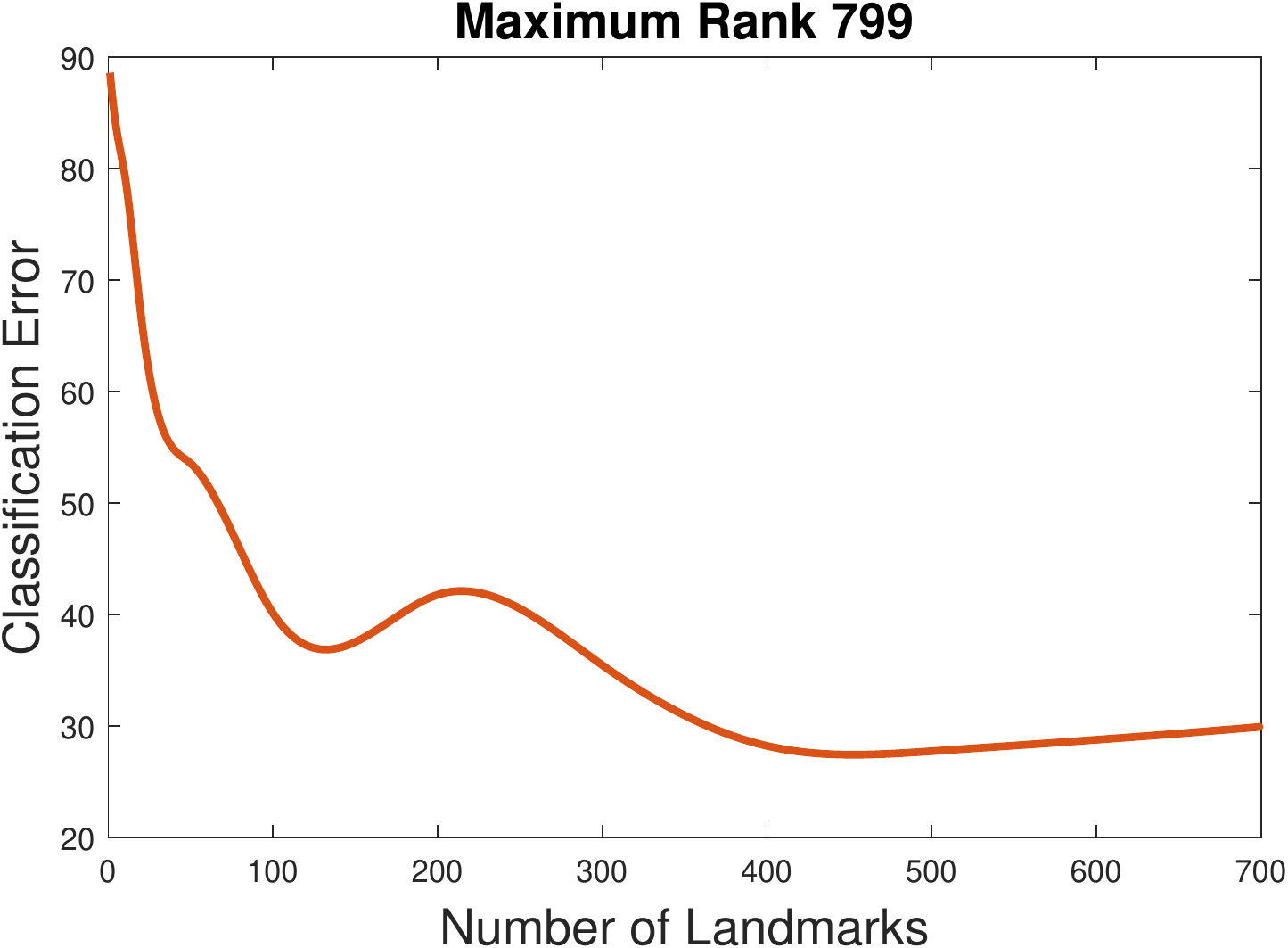}
        \subcaption{Office - Caltech \label{FigImageEL}}
    \end{subfigure}
    \begin{subfigure}[c]{0.49\textwidth}
        \includegraphics[width=1\textwidth]{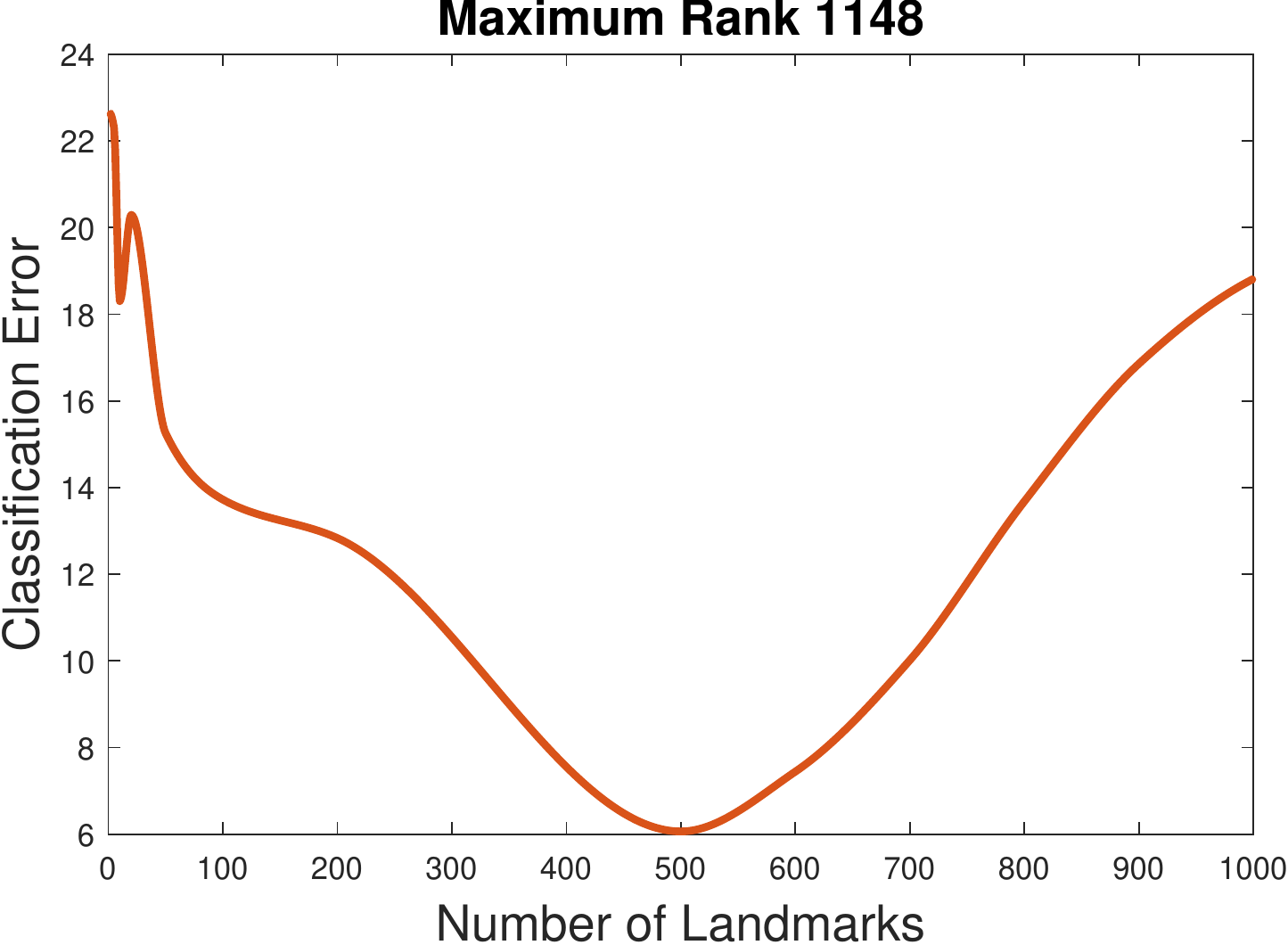}
        \subcaption{Reuters\label{FigTextEL}}
    \end{subfigure}
    \caption{Relationship between number of landmarks and classification error as mean over all Office vs Caltech datasets, shown in left figure, and mean over all Reuters datasets, shown in right figure. Number of landmarks for datasets smaller as maximum rank are determined via $min({n,d})$. \label{FigErrorvslandmarks}}
\end{figure}

The sensitivity of the number of landmarks on prediction error as the only parameter of \ac{NBT} is demonstrated in figure~\ref{FigErrorvslandmarks}.
It shows a comparison of the number of landmarks and the mean of classification error over \textit{Reuters} and \textit{Office - Caltech} datasets.
The plot indicates that the error decreases to a global minimum with an increasing number of landmarks, which supports the Nyström error expectation.
However, further increasing to the maximum number of landmarks, i.e.\ all samples, the error starts to increase.
We assume this indicates that only a subset of features is relevant for classification and remaining features are noise.
Further, this subset is drawn randomly, hence by choosing various landmark matrices, other features becoming relevant or non-relevant, as features correlate with certain features and again other features correlate with others.

\subsection{Comparison of Model Complexity}\label{sec:complexity}
\begin{table}[b]
    \centering
    \resizebox{\textwidth}{!}{%
        \begin{tabular}{c c c c c c c c c c }
            \toprule
            Dataset           & \ac{PCVM}&\ac{SVM} & \ac{TCA}   & \ac{JDA}  &\ac{SA}  & \ac{TKL} & PCTKVM  & NTVM   \\ \toprule
            Reuters(1153.66)      &    49.07  & 441.78       & 168.51&201.87&     100.87& 351.21   &\textbf{1.97}    &329.51                                             \\
            Image(633.25)         &    62.87   & 284.37       &    231.65    &264.38     &238.44&    262.64  &    46.63 &\textbf{27.59}  \\
            20 Newsgroup(3758.30) &\textbf{74.23} & 1247.10          &269.75& 252.49 & 211.57 & 1046.26     &    92.89& 74.70    \\ \bottomrule
            Overall Mean          & 62.06     & 640.17          & 223.30 &  245.31 & 183.60 & 553.27 &\textbf{ 47.16} & 143.93
        \end{tabular}}
        \caption{Average mean number of model vectors of a classifier for Reuters, Image and Newsgroup datasets. The average number of examples in the datasets are shown on the right side of the dataset name in brackets.\label{TableMeanNSV}}
\end{table}

We measured the model complexity with the number of model vectors, e.g.\ support vectors.
The result of our experiment is shown as mean summarizing a dataset group in table~\ref{TableMeanNSV}. We see that the transfer learning models of the {PCTKVM} provide very sparse models while having good performance.
The sparsity of \ac{NTVM} is also very competitive. However, the overall sparsity is worse in comparison to \ac{PCTKVM} and \ac{PCVM}.
In comparison to all non-\ac{PCVM} methods, the \ac{PCTKVM} outperforms the respective methods by far.

\begin{figure}[]
    \caption{Sample run on \textit{Orgs vs People} (Text dataset). Red colors for the class  \textit{Orgs} and blue for the class \textit{People}. This plot includes training and testing data. Model complexity of {PCTKVM} on the left and {\ac{SVM}} on the right. The {PCTKVM} uses three vectors and achieves an error of $21\%$. The {\ac{TKL}-\ac{SVM}} needs $339$ vectors and has an error of $19\%$. The black circled points are used model vectors. Reduced with t-SNE \cite{vanDerMaaten.2008}. Best viewed in color.\label{FigModel}}
    \begin{subfigure}[c]{0.49\textwidth}
        \includegraphics[width=1\textwidth]{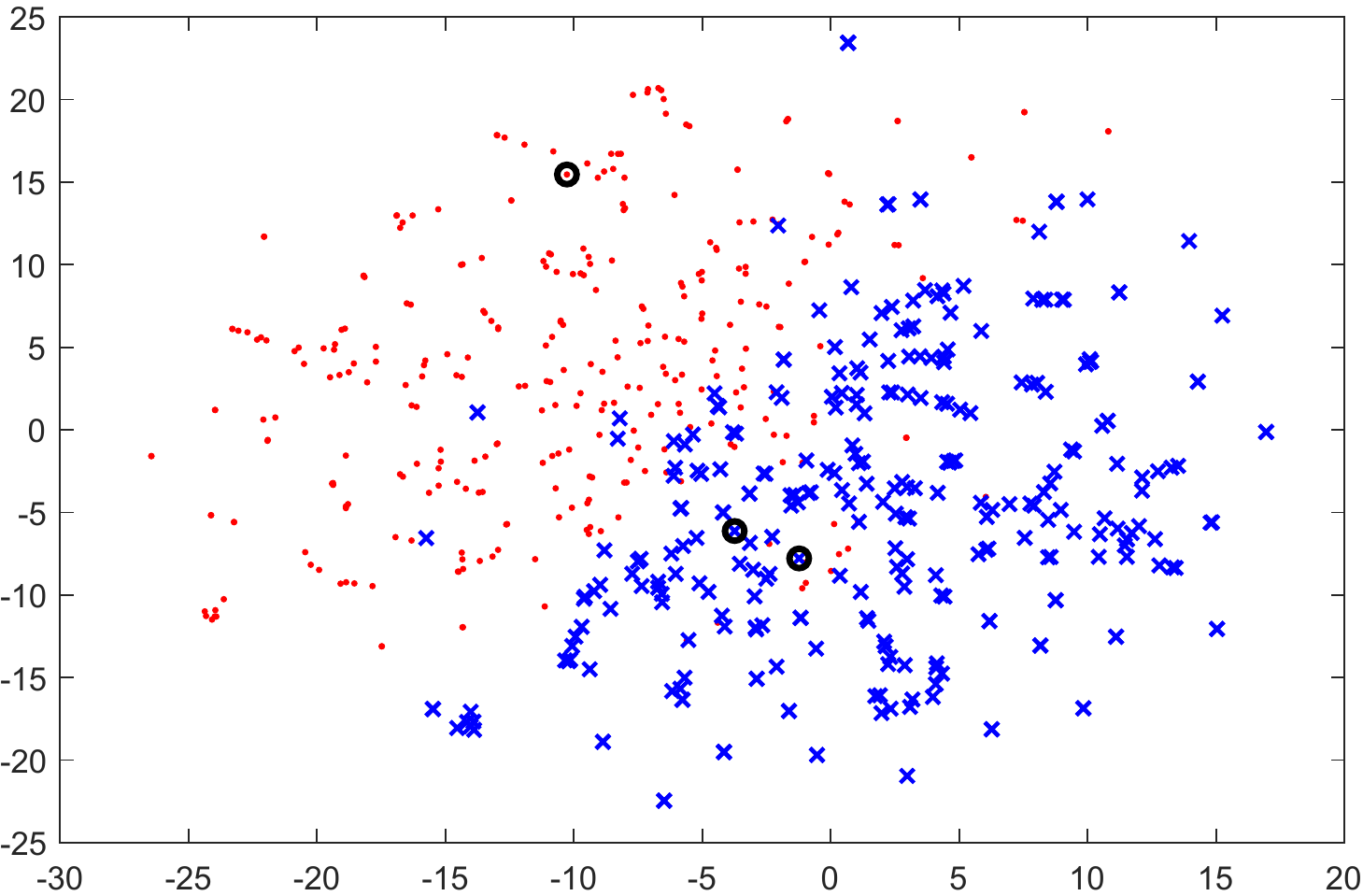}
        \caption{PCTKVM \label{FigModelPCVM}}
    \end{subfigure}
      \begin{subfigure}[c]{0.49\textwidth}
        \includegraphics[width=1\textwidth]{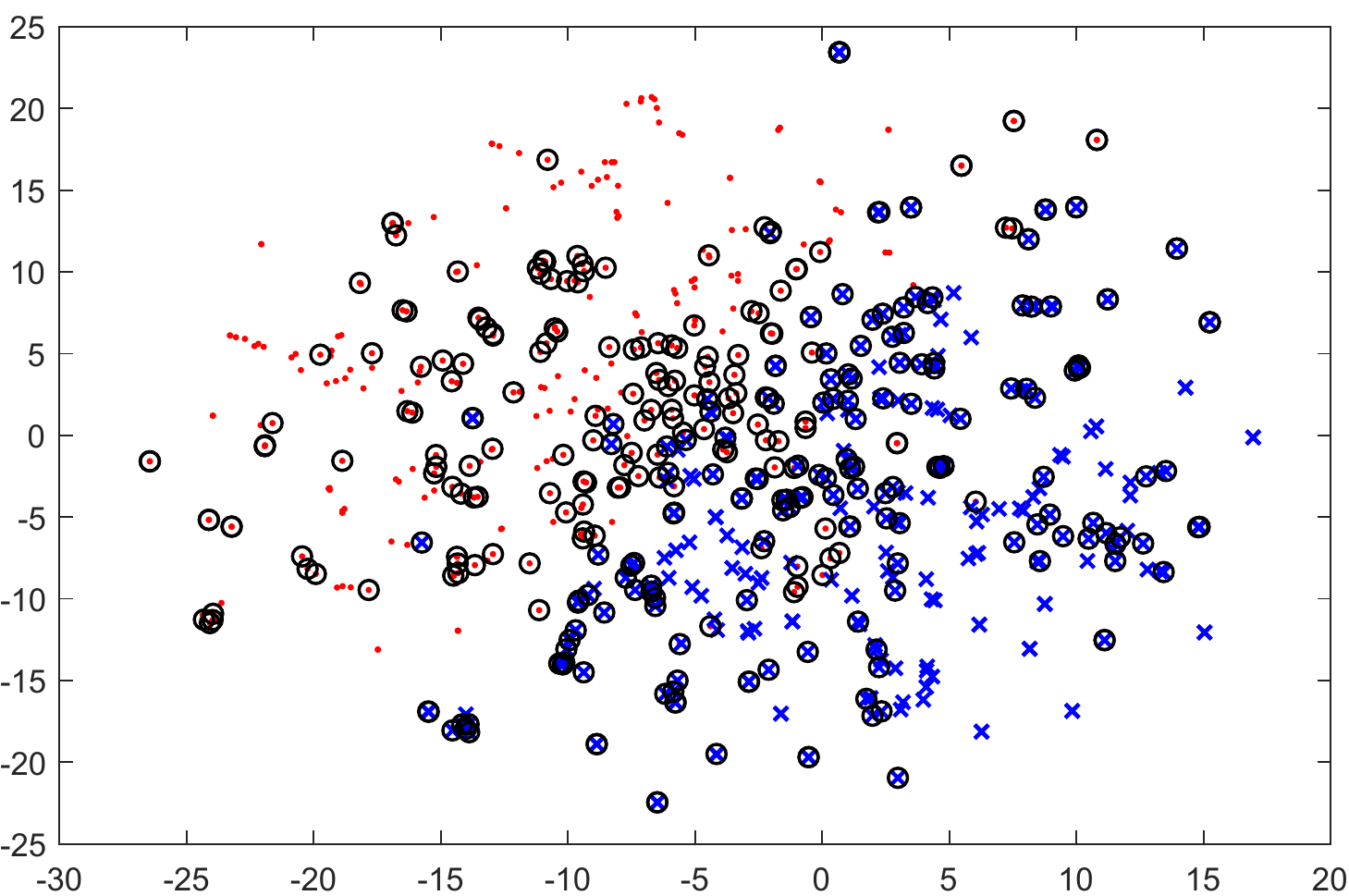}
        \caption{TKL-SVM \label{FigModelSVM}}
    \end{subfigure}
\end{figure}

The difference in model complexity is exemplarily shown in figure~\ref{FigModel}.
It shows a sample result of classification of {PCTKVM} and {\ac{TKL}-\ac{SVM}} on the text dataset \textit{Orgs vs People} with the settings from above.
The PCTKVM error is $21\%$ with three model vectors and the error of \ac{TKL}-\ac{SVM} is $19\%$ with $339$ support vectors.\\
PCTKVM achieves sustain performance by a small model complexity and provides a way to interpret the model.
Note that the algorithms are trained in the original feature space and the models are plotted in a reduced space,
using the t-distributed stochastic neighbor embedding algorithm \cite{vanDerMaaten.2008}. Note that the Kullback-Leibler divergence of the data shown in figure~\ref{FigModel} between input distribution (original space) and output distribution (reduced space) is 0.92.

\subsection{Time Comparison}
The mean time results in seconds of the cross validation study per data set group are shown in table~\ref{tab:result_time}. Note that \ac{SVM}  and \ac{PCVM} are underlying classifier for compared approaches and are presented for the baseline and not marked as winners in the table. They are also included in the time measurement of transfer learning approaches. Overall the \ac{SVM} is the fastest algorithm, because it is baseline and uses the LibSVM implementation. The overall fastest transfer learning approach is \ac{TKL}, but \ac{JDA} is also promising and fastest at Reuters and Newsgroup.\newline
The PCVM is overall by far the slowest classifier. By integration of \ac{TKL} and \ac{NBT} the time performance of resulting \ac{PCTKVM} and \ac{NTVM} are a big magnitude faster. We assume that the \ac{PCVM} converges faster with transfer learning resulting in less computational time. Overall \ac{NTVM} is slightly faster and lower in time at Reuters and Newsgroup in comparison to \ac{PCTKVM}, which supports the discussion about computational complexity in section~\ref{sec:props_nbt}. In comparison to other transfer learning approaches, both approaches are slower than other transfer learning approaches. But the reason for this should be the \ac{PCVM} as underlying classifier, because \ac{TKL} is the fastest transfer approach with \ac{SVM}. Further work should aim to measure the time with same classifier to make results more comparable.
\begin{table}[]%
     \resizebox{\textwidth}{!}{%
\begin{tabular}{lccccccccc}
    \toprule
    Dataset    & \specialcell{SVM \\Baseline}   & TCA   & JDA    & TKL   & \textbf{}SA && \specialcell{PCVM \\Baseline}& PCTKVM& NTVM \\\midrule
    Reuters    &    0.06  &  0.86 &  0.36&  0.40 &  0.87   &             & 543.71     &     16.20 &  8.78 \\
    Newsgroup  &    1.35  &    21.39&    {4.79}&   2.80&   59.70 &         & 1501.7     &     5.06  & 25.29\\
    Image      &    0.02  &    0.29&    0.16 &    0.44 &      {0.08} &      & 258.78     &    17.69 &    3.41 \\\midrule
    Overall    &    0.48 & 7.51 & 1.77 & {1.21} &  20.22 &                            & 768.06     &     12.98 &  12.49\\
\end{tabular}}
\caption{Result of cross-validation test shown in mean time in seconds per dataset group. The baseline classifiers are \ac{SVM} is underlying classifier of \ac{TCA}, \ac{JDA}, \ac{TKL} and \ac{SA} and the \ac{PCVM} is underlying classifier of \ac{PCTKVM} and \ac{NTVM}. Note that the \ac{SVM} time is naturally lower to transfer learning approaches.\label{tab:result_time}}
\end{table}
\FloatBarrier
\section{Conclusion}\label{sec:conclusion}
Summarizing, we proposed two transfer learning extensions for the \ac{PCVM}, resulting in \ac{PCTKVM} and \ac{NTVM}. The first shows the best overall sparsity and comparable performance to common transfer learning approaches. The \ac{NTVM} has an outstanding performance, both in absolute values and statistical significance. It has competitive sparsity and lowest computational complexity compared to discussed solutions. \ac{NBT} is an enhancement of previous versions of Basis Transfer via Nyström methods and is no longer limited to specific domain adaptation tasks.
The dimensionality reduction paired with projection of source data into the target subspace via \ac{NBT} showed its reliability and robustness in this study.
Proposed solutions are tested against standard benchmarks in the field in terms of algorithms and datasets. In future work, deep transfer learning, different baseline  classifiers and real-world or different domain adaptation datasets should be integrated. Further, smart sampling techniques for landmark selection should be tackled.




\bibliographystyle{elsarticle-num}
\bibliography{nc_raab_schleif_19}

\end{document}